# Identification of Conditional Interventional Distributions

**Ilya Shpitser and Judea Pearl**
Cognitive Systems Laboratory
Department of Computer Science
University of California, Los Angeles
Los Angeles, CA. 90095
$\{ilyas, judea\}@cs.ucla.edu$

## Abstract

The subject of this paper is the elucidation of effects of actions from causal assumptions represented as a directed graph, and statistical knowledge given as a probability distribution. In particular, we are interested in predicting distributions on post-action outcomes given a set of measurements. We provide a necessary and sufficient graphical condition for the cases where such distributions can be uniquely computed from the available information, as well as an algorithm which performs this computation whenever the condition holds. Furthermore, we use our results to prove completeness of $do$-calculus [Pearl, 1995] for the same identification problem, and show applications to sequential decision making.

## 1 Introduction

This paper deals with computing effects of actions in domains represented as $causal\ diagrams$, or graphs with directed and bidirected arcs. Such diagrams contain vertices corresponding to variables of interest, directed arcs representing potential direct causal relationships, and bidirected arcs which are spurious correlations or 'hidden common causes' [Pearl, 1995], [Pearl, 2000]. Aside from this kind of causal knowledge represented by the graph, we also have statistical knowledge about the variables in the model, represented by a joint probability distribution $P$.

An $action$ on a set of variables $\mathbf{X}$ consists of forcing the variables to particular values $\mathbf{x}$ regardless of the value $\mathbf{X}$ would have otherwise attained. This action, denoted by $do(\mathbf{x})$ in [Pearl, 2000], transforms the original distribution $P$ into an $interventional\ distribution$ denoted by $P_{\mathbf{x}}$. We quantify the effect of the action $do(\mathbf{x})$ on a set $\mathbf{Y}$ by considering $P_{\mathbf{x}}(\mathbf{y})$. In this paper, we also consider conditional effects of the form $P_{\mathbf{x}}(\mathbf{y}|\mathbf{z})$, which correspond to the effect of $do(\mathbf{x})$ on $\mathbf{Y}$ in a situation where it is known that $\mathbf{z}$ holds.

The problem of $causal\ effect\ identifiability$ consists of finding graphs in which effects represented by $P_{\mathbf{x}}(\mathbf{y})$ or $P_{\mathbf{x}}(\mathbf{y}|\mathbf{z})$ can be uniquely determined from the original distribution $P$. It is well known that in causal diagrams with no bidirected arcs, corresponding to Markovian models, all effects are identifiable [Pearl, 2000]. The situation is more complicated in causal diagrams containing bidirected arcs, and the corresponding models which are called semi-Markovian. Consider the graphs in Fig. 1. Here $P_x(y)$ is not identifiable in $G$ in Fig. 1 (a), but identifiable and equal to $P(y)$ in $G'$ in Fig. 1 (b).

Conditioning can both help and hinder identifiability. In the graph $G$, conditioning on $Z$ renders $Y$ independent of any changes to $X$, making $P_x(y|z)$ equal to $P(y|z)$. On the other hand, in $G'$, conditioning on $Z$ makes $X$ and $Y$ dependent, resulting in $P_x(y|z)$ becoming non-identifiable.

The past decade has yielded several sufficient conditions for identifiability in the semi-Markovian case [Spirtes, Glymour, & Scheines, 1993], [Pearl & Robins, 1995], [Pearl, 1995], [Kuroki & Miyakawa, 1999]. An overview of this work can be found in [Pearl, 2000]. Generally, sufficiency results for this problem rely on the causal and probabilistic assumptions embedded in the graph, and are phrased as graphical criteria. For example, it is known that whenever a set $\mathbf{Z}$ of non-descendants of $\mathbf{X}$ blocks certain paths in the graph from $\mathbf{X}$ to $\mathbf{Y}$, then $P_{\mathbf{x}}(\mathbf{y}) = \sum_{\mathbf{z}} P(\mathbf{y}|\mathbf{z},\mathbf{x})P(\mathbf{z})$ [Pearl, 2000].

Identification of causal effects can also be deduced by algebraic methods. [Pearl, 1995] provided 3 rules of $do$-calculus, which systematically use properties of the graph to manipulate interventional distribution expressions. These manipulations can be applied until the effect is reduced to something computable from $P$. Similarly, [Halpern, 2000] constructed a system of axioms and inference rules which can frame the identification problem as one of theorem proving. The axiom set was then shown to be complete. Algebraic methods such as these have the disadvantage of requiring the user to come up with a proof strategy to show identifiability in any given case, rather than giving an explicit graphical criterion.

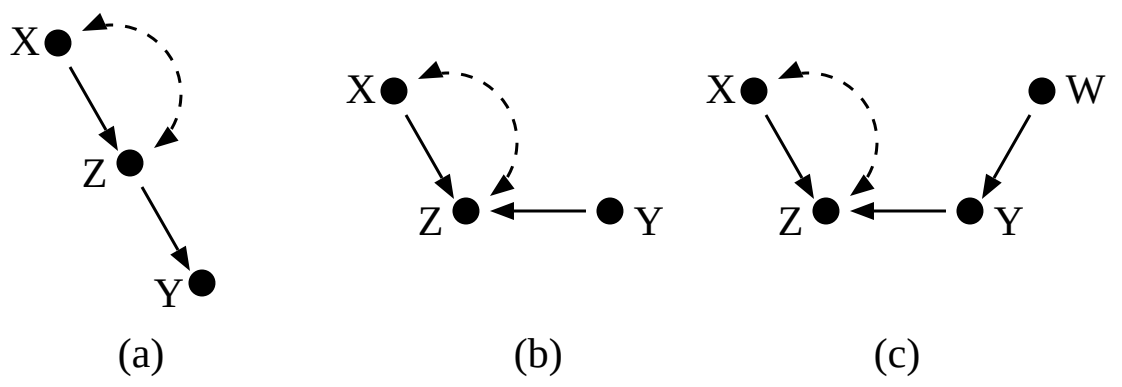

Figure 1: (a) Graph $G$. (b) Graph $G'$. (c) Graph $G''$.

A number of necessity results have recently been derived as well. One such result [Tian & Pearl, 2002] states that $P_x$ is identifiable if and only if there is no path consisting entirely of bidirected arcs from $X$ to a child of $X$. A recent paper [Shpitser & Pearl, 2006] constructed a complete algorithm for identifying $P_{\mathbf{x}}(\mathbf{y})$, and showed $do$-calculus and a version of Tian's algorithm [Tian, 2002] to be complete for the same identification problem. The last result is also shown in [Huang & Valtorta, 2006]. A general algorithm for identifying $P_{\mathbf{x}}(\mathbf{y}|\mathbf{z})$ has been proposed in [Tian, 2004]. Unfortunately, as we will later show, the algorithm has its limitations.

In this paper, we use the above necessity results to solve the problem of identifying $conditional$ distributions $P_{\mathbf{x}}(\mathbf{y}|\mathbf{z})$. We show a way to reduce this problem to identifying $unconditional$ distributions of the form $P_{\mathbf{x}'}(\mathbf{y}')$, for which complete criteria and algorithms are already known. We then use this reduction to give a complete graphical criterion for identification of conditional effects, and a complete algorithm which returns an expression for such an effect in terms of $P$ whenever the criterion holds. We use our results to prove completeness of $do$-calculus [Pearl, 1995] for identifying conditional effects.

## 2 Motivation: Sequential Decisions

Our interest in conditional interventional distributions can be motivated by their relationship to sequential decision problems that arise in many domains [Pearl & Robins, 1995]. We will use an example from treatment management in medicine. A patient comes in, complaining of a set of symptoms $\mathbf{Z}^0$. After administering some tests $\mathbf{Z}^1$ of his own, the doctor prescribes a treatment $X^1 = g_1(\mathbf{Z}^0, \mathbf{Z}^1)$. After a time, additional tests $\mathbf{Z}^2$ are done to check for medication side effects, patient improvement, complications, and so on. Possibly a followup treatment $X^2 = g_2(\mathbf{Z}^2, X^1)$ is prescribed. In general, the treatment $X^i$ at time $i$ is a function of treatment history $\mathbf{X}^{<i}$ and observation history $\mathbf{Z}^{<i}$. The treatment process continues until the patient gets well or dies – represented by the state of $outcome$ $variables$ $\mathbf{Y}$. Note that in this situation, the doctor performs interventions $do(x^i)$, but the specific values of the treatment variables are not known in advance, but instead depend on symptoms and test results performed 'on the fly' via policy functions $g_i$.

The effect of this kind of conditional policy $\mathbf{g} = \{g_i | i = 1, ..., K\}$ is represented as the probability distribution on $\mathbf{Y}$ given that the treatment variables $\mathbf{X} = \{X^i | i = 1, ..., K\}$ are fixed according to $\mathbf{g}$. We write this is $P(\mathbf{Y}_{\mathbf{X}_\mathbf{g}})$. To evaluate the efficacy of $\mathbf{g}$, it would be useful to determine $P(\mathbf{Y}_{\mathbf{X}_\mathbf{g}})$ from statistical data available to the hospital regarding similar cases in the past, rather than resorting to testing the policy on the patient. Note that this effect is a complex counterfactual quantity because observation history $\mathbf{Z}^{<i}$ and treatment history $\mathbf{X}^{<i}$ are mutually dependent (by either $\mathbf{g}$, or the normal causative mechanisms in the model). Nevertheless it can be shown that by doing case analysis on $\mathbf{Z} = \bigcup_i \mathbf{Z}^i$, we can obtain $P(\mathbf{y}_{\mathbf{X}_\mathbf{g}}) = \sum_{\mathbf{z}} P_{\mathbf{x}_\mathbf{z}}(\mathbf{y}|\mathbf{z}) P_{\mathbf{x}_\mathbf{z}}(\mathbf{z})$, where $\mathbf{x}_\mathbf{z}$ is the treatment prescribed given that $\mathbf{Z}$ is observed to be $\mathbf{z}$. The key observation is that since the policy $\mathbf{g}$ is known in advance, fixing $\mathbf{z}$ determines $\mathbf{x}_\mathbf{z}$ in advance as well. Thus, the effect of sequential conditional plans can be identified from data if we can find a way of identifying conditional distributions of the form $P_{\mathbf{x}}(\mathbf{y}|\mathbf{z})$.

## 3 Notation and Definitions

In this section we reproduce the technical definitions needed for the rest of the paper. We will denote variables by capital letters, and their values by small letters. Similarly, sets of variables will be denoted by bold capital letters, and sets of values by bold small letters. We will use some graph-theoretic abbreviations: $Pa(\mathbf{Y})_G$, $An(\mathbf{Y})_G$, and $De(\mathbf{Y})_G$ will denote the set of (observable) parents, ancestors, and descendants of the node set $\mathbf{Y}$ in $G$, respectively. We will omit the graph subscript if the graph in question is assumed or obvious. We will denote the set $\{X \in G | De(X)_G = \emptyset\}$ as the root set of $G$. Finally, following [Pearl, 2000], we will denote $G_{\overline{\mathbf{x}}\underline{\mathbf{y}}}$ to be an edge subgraph of $G$ where all incoming arrows into $\mathbf{X}$ and all outgoing arrows from $\mathbf{Y}$ are deleted.

Having fixed our notation, we can proceed to formalize the notions discussed in the previous section. A probabilistic causal model is a tuple $M = \langle \mathbf{U}, \mathbf{V}, \mathbf{F}, P(\mathbf{U}) \rangle$, where $\mathbf{V}$ is a set of observable variables, $\mathbf{U}$ is a set of unobservable variables distributed according to $P(\mathbf{U})$, and $\mathbf{F}$ is a set of functions. Each variable $V \in \mathbf{V}$ has a corresponding function $f_V \in \mathbf{F}$ that determines the value of $V$ in terms of other variables in $\mathbf{V}$ and $\mathbf{U}$. The distribution on $\mathbf{V}$ induced by $P(\mathbf{U})$ and $\mathbf{F}$ will be denoted $P(\mathbf{V})$.

Sometimes it is assumed $P(\mathbf{V})$ is a positive distribution. In this paper we do not make this assumption. Thus, we must make sure that for every distribution $P(\mathbf{W}|\mathbf{Z})$ that we consider, $P(\mathbf{Z})$ is positive. This can be achieved by making sure to sum over events with positive probability only, and by only considering observations with positive probability. Furthermore, for any action $do(\mathbf{x})$ that we consider, it must be the case that $P(\mathbf{x}|Pa(\mathbf{X})_G \setminus \mathbf{X}) > 0$ otherwise the distribution $P_{\mathbf{x}}(\mathbf{V})$ is not well defined [Pearl, 2000]. Finally, because $P_{\mathbf{x}}(\mathbf{y}|\mathbf{z}) = P_{\mathbf{x}}(\mathbf{y},\mathbf{z})/P_{\mathbf{x}}(\mathbf{z})$, we must make sure that $P_{\mathbf{x}}(\mathbf{z}) > 0$.

The induced graph $G$ of a causal model $M$ contains a node for every element in $\mathbf{V}$, a directed edge between nodes $X$ and $Y$ if $f_Y$ possibly uses the values of $X$ directly to determine the value of $Y$, and a bidirected edge between nodes $X$ and $Y$ if $f_X$ and $f_Y$ both possibly use the value of some variable in $\mathbf{U}$ to determine their respective values. In this paper we consider $recursive$ causal models, those models which induce acyclic graphs.

In any causal model there is a relationship between its induced graph $G$ and $P$, where $P(x_1, ..., x_n) = \prod_i P(x_i|pa^*(X_i)_G)$, and $Pa^*(.)_G$ also includes unobservable parents [Pearl, 2000]. Whenever this relationship holds, we say that $G$ is an I-map (independence map) of $P$. The I-map relationship allows us to link independence properties of $P$ to $G$ by using the following well-known notion of path separation [Pearl, 1988].

**Definition 1 (d-separation)** *A path $p$ in $G$ is said to be d-separated by a set $\mathbf{Z}$ if and only if either*

1 *$p$ contains a chain $I \to M \to J$ or fork $I \leftarrow M \to J$, such that $M \in \mathbf{Z}$, or*

2 *$p$ contains an inverted fork $I \to M \leftarrow J$ such that $De(M)_G \cap \mathbf{Z} = \emptyset$.*

Two sets $\mathbf{X}, \mathbf{Y}$ are said to be d-separated given $\mathbf{Z}$ in $G$ if all paths from $\mathbf{X}$ to $\mathbf{Y}$ in $G$ are d-separated by $\mathbf{Z}$. The following well-known theorem links d-separation of vertex sets in an I-map $G$ with the independence of corresponding variable sets in $P$.

**Theorem 1** *If sets $\mathbf{X}$ and $\mathbf{Y}$ are d-separated by $\mathbf{Z}$ in $G$, then $\mathbf{X} \perp\!\!\!\perp \mathbf{Y}|\mathbf{Z}$ in every $P$ for which $G$ is an I-map.*

A path that is not d-separated is said to be $d-connected$. A d-connected path starting from a node $X$ with an arrow pointing to $X$ is called a $back-door\ path$ from $X$.

In the framework of causal models, actions are modifications of functional relationships. Each action $do(\mathbf{x})$ on a causal model $M$ produces a new model $M_{\mathbf{x}} = \langle \mathbf{U}, \mathbf{V}, \mathbf{F}_{\mathbf{x}}, P(\mathbf{U})\rangle$, where $\mathbf{F}_{\mathbf{x}}$, is obtained by replacing $f_X \in \mathbf{F}$ for every $X \in \mathbf{X}$ with a new function that outputs a constant value $x$ given by $do(\mathbf{x})$. Note that $M_{\mathbf{x}}$ induces $G \setminus \mathbf{X}$. Since subscripts are used to denote submodels, we will use numeric superscripts to enumerate models and their associated probability distributions (e.g. $M^1, P^1$).

We can now define formally the notion of identifiability of conditional effects from observational distributions.

**Definition 2 (Causal Effect Identifiability)** *The causal effect of an action $do(\mathbf{x})$ on a set of variables $\mathbf{Y}$ in a given context $\mathbf{z}$ such that $\mathbf{Y}, \mathbf{X}, \mathbf{Z}$ are disjoint is said to be identifiable from $P$ in $G$ if $P_{\mathbf{x}}(\mathbf{y}|\mathbf{z})$ is (uniquely) computable from $P$ in any causal model which induces $G$.*

Note that because $\mathbf{Z}$ can be empty, conditional effects generalize the more commonly used notion of effect as $P_{\mathbf{x}}(\mathbf{y})$. The following lemma establishes the conventional technique used to prove non-identifiability in a given $G$.

**Lemma 1** *Let $\mathbf{X}, \mathbf{Y}, \mathbf{Z}$ be sets of variables. Assume there exist two causal models $M^1$ and $M^2$ with the same induced graph $G$ such that $P^1(\mathbf{V}) = P^2(\mathbf{V})$, $P^1(\mathbf{x}|Pa(\mathbf{X})_G \setminus \mathbf{X}) > 0$, $P^1_{\mathbf{x}}(\mathbf{z}) > 0$, $P^2_{\mathbf{x}}(\mathbf{z}) > 0$, and $P^1_{\mathbf{x}}(\mathbf{Y}|\mathbf{z}) \neq P^2_{\mathbf{x}}(\mathbf{Y}|\mathbf{z})$. Then $P_{\mathbf{x}}(\mathbf{y}|\mathbf{z})$ is not identifiable in $G$.*

Throughout the paper, we will make use of 3 rules of $do$-calculus [Pearl, 1995]. These identities, derived from Theorem 1, are known to hold for interventional distributions:

Rule 1: $P_{\mathbf{x}}(\mathbf{y}|\mathbf{z},\mathbf{w}) = P_{\mathbf{x}}(\mathbf{y}|\mathbf{w})$ if $(\mathbf{Y} \perp\!\!\!\perp \mathbf{Z}|\mathbf{X},\mathbf{W})_{G_{\overline{\mathbf{X}}}}$

Rule 2: $P_{\mathbf{x},\mathbf{z}}(\mathbf{y}|\mathbf{w}) = P_{\mathbf{x}}(\mathbf{y}|\mathbf{z},\mathbf{w})$ if $(\mathbf{Y} \perp\!\!\!\perp \mathbf{Z}|\mathbf{X},\mathbf{W})_{G_{\overline{\mathbf{X}},\underline{\mathbf{Z}}}}$

Rule 3: $P_{\mathbf{x},\mathbf{z}}(\mathbf{y}|\mathbf{w}) = P_{\mathbf{x}}(\mathbf{y}|\mathbf{w})$ if $(\mathbf{Y} \perp\!\!\!\perp \mathbf{Z}|\mathbf{X},\mathbf{W})_{G_{\overline{\mathbf{X}},\overline{\mathbf{Z}^*}}}$

where $\mathbf{Z}^* = \mathbf{Z} \setminus An(\mathbf{W})_{G_{\overline{\mathbf{X}}}}$.

Note that one way to restate rule 2 is to say that it holds for a set $\mathbf{Z}$ whenever there are no back-door paths from $\mathbf{Z}$ to $\mathbf{Y}$ given the action $do(\mathbf{x})$ and observations $\mathbf{w}$.

## 4 Hedges and Unconditional Effects

A previous paper [Shpitser & Pearl, 2006] provided a complete algorithm, and a corresponding graphical condition for identification of all effects of the form $P_{\mathbf{x}}(\mathbf{y})$. In this section, we will provide an overview of these results. We first consider a set of nodes mutually interconnected by bidirected arcs.

**Definition 3 (C-component)** *Let $G$ be a semi-Markovian graph such that a subset of its bidirected arcs forms a spanning tree over all vertices in $G$. Then $G$ is a C-component (confounded component).*

Any causal diagram is either a C-component, or can be uniquely partitioned into a set $C(G)$ of maximal C-components. C-components are an important notion for identifiability and were studied at length in [Tian, 2002]. The importance of this structure stems from the fact that the distribution $P$ factorizes into a set of interventional distributions according to $C(G)$. For example the graph in Fig. 1 (a) has two C-components, $\{X, Z\}$ and $\{Y\}$, so $P(x, y, z) = P_{x,z}(y)P_y(x, z)$. Furthermore, each term function **ID**($\mathbf{y}$, $\mathbf{x}$, P, G)
INPUT: $\mathbf{x}$,$\mathbf{y}$ value assignments, P a probability distribution, G a causal diagram (an I-map of P).
OUTPUT: Expression for $P_{\mathbf{x}}(\mathbf{y})$ in terms of P or **FAIL**(F,F').

1 if $\mathbf{x} = \emptyset$, return $\sum_{\mathbf{v} \setminus \mathbf{y}} P(\mathbf{v})$.

2 if $\mathbf{V} \neq An(\mathbf{Y})_G$,
return $\mathbf{ID}(\mathbf{y}, \mathbf{x} \cap An(\mathbf{Y})_G, P(An(\mathbf{Y})), An(\mathbf{Y})_G)$.

3 let $\mathbf{W} = (\mathbf{V} \setminus \mathbf{X}) \setminus An(\mathbf{Y})_{G_{\overline{\mathbf{x}}}}$.
if $\mathbf{W} \neq \emptyset$, return $\mathbf{ID}(\mathbf{y}, \mathbf{x} \cup \mathbf{w}, P, G)$.

4 if $C(G \setminus \mathbf{X}) = \{S_1, ..., S_k\}$,
return $\sum_{\mathbf{v} \setminus (\mathbf{y} \cup \mathbf{x})} \prod_i \mathbf{ID}(s_i, \mathbf{v} \setminus s_i, P, G)$.

if $C(G \setminus \mathbf{X}) = \{S\}$,

5 if $C(G) = \{G\}$, throw $\mathbf{FAIL}(G, S)$.

6 if $S \in C(G)$, return $\sum_{s \setminus \mathbf{y}} \prod_{V_i \in S} P(v_i | v_\pi^{(i-1)})$.

7 if $(\exists S') S \subset S' \in C(G)$, return $\mathbf{ID}(\mathbf{y}, \mathbf{x} \cap S',$
$\prod_{V_i \in S'} P(V_i | V_\pi^{(i-1)} \cap S', v_\pi^{(i-1)} \setminus S'), S')$.

Figure 2: A complete identification algorithm. **FAIL** propagates through recursive calls like an exception, and returns $F, F'$ which form the hedge which witnesses non-identifiability of $P_{\mathbf{x}}(\mathbf{y})$. $\pi$ is some topological ordering of nodes in $G$.

in this factorization is identifiable. These observations allow one to decompose the identification problem into a set of subproblems. We use C-components to define a graph structure which turns out to be a key presence in all unidentifiable effects.

**Definition 4 (C-forest)** *Let $G$ be a semi-Markovian graph, where **Y** is the root set. Then $G$ is a **Y**-rooted $C$-$forest$ if all nodes in $G$ form a C-component, and all observable nodes have at most one child.*

If two C-forests share the same root set, and only one of them contains any nodes in $\mathbf{X}$, then the resulting graph structure witnesses the non-identifiability of certain effects of $do(\mathbf{x})$. The structure in question is called a hedge.

**Definition 5 (hedge)** *Let $\mathbf{X}, \mathbf{Y}$ be sets of variables in $G$. Let $F, F'$ be **R**-rooted C-forests such that $F \cap \mathbf{X} \neq \emptyset$, $F' \cap \mathbf{X} = \emptyset$, $F' \subseteq F$, and $\mathbf{R} \subset An(\mathbf{Y})_{G_{\overline{\mathbf{x}}}}$. Then $F$ and $F'$ form a hedge for $P_{\mathbf{x}}(\mathbf{y})$.*

Hedges precisely characterize non-identifiability of interventional joint distributions, as the following results show.

**Theorem 2** *Assume there exist **R**-rooted C-forests $F, F'$ that form a hedge for $P_{\mathbf{x}}(\mathbf{y})$ in $G$. Then $P_{\mathbf{x}}(\mathbf{y})$ is not identifiable in $G$.*

*Proof:* Consider the graph $H = An(\mathbf{Y})_G \cap De(F)_G$, and two models $M^1, M^2$ which induce $H$. All variables in both models are binary. In $M^1$ every variable is equal to the bit parity of its parents. In $M^2$ the same is true, except all nodes in $F'$ disregard the parent values in $F \setminus F'$. All **U** are fair coins in both models. It has been shown in [Shpitser & Pearl, 2006] that $M^1$ and $M^2$ satisfy the conditions in Lemma 1 for $P_{\mathbf{x}}(\mathbf{y})$. □

**Theorem 3 (hedge criterion)** *$P_{\mathbf{x}}(\mathbf{y})$ is identifiable from $P$ in $G$ if and only if there does not exist a hedge for $P_{\mathbf{x}'}(\mathbf{y}')$ in $G$, for any $\mathbf{X}' \subseteq \mathbf{X}$ and $\mathbf{Y}' \subseteq \mathbf{Y}$.*

The proof can be found in [Shpitser & Pearl, 2006]. Whenever $P_{\mathbf{x}}(\mathbf{y})$ is identifiable, we say that $P_{\mathbf{x}}(\mathbf{y})$ does not contain any hedges. In such a case, the **ID** algorithm (pictured in Fig. 2) computes the expression for $P_{\mathbf{x}}(\mathbf{y})$ in terms of $P$. It has also been shown in [Shpitser & Pearl, 2006] that whenever $P_{\mathbf{x}}(\mathbf{y})$ is not identifiable, **ID** returns the witnessing hedge, which entails the following result.

**Theorem 4** ***ID** is complete.*

The previous results were also used to derive a completeness result for $do$-calculus

**Theorem 5** *The rules of $do$-calculus, together with standard probability manipulations are complete for determining identifiability of all effects of the form $P_{\mathbf{x}}(\mathbf{y})$.*

*Proof:* The proof, which reduces steps of the **ID** algorithm to sequences of applications of $do$-calculus, can be found in [Shpitser & Pearl, 2006]. □

Armed with a complete criterion and corresponding algorithm for identifying $P_{\mathbf{x}}(\mathbf{y})$, we tackle the conditional version of the problem in the next section.

## 5 Identifiability of Conditional Interventional Distributions

We now consider the problem of identifying distributions of the form $P_{\mathbf{x}}(\mathbf{y}|\mathbf{w})$, where $\mathbf{X}, \mathbf{Y}, \mathbf{W}$ are arbitrary disjoint sets of variables. Our approach will be to reduce this problem to a solved case where the set $\mathbf{W}$ is empty. One way to accomplish this is to use rule 2 of $do$-calculus to transform conditioning on $\mathbf{W}$ into interventions. Recall that whenever rule 2 applies to a set $\mathbf{Z} \subseteq \mathbf{W}$ in $G$ for $P_{\mathbf{x}}(\mathbf{y}|\mathbf{w})$ then $P_{\mathbf{x}}(\mathbf{y}|\mathbf{w}) = P_{\mathbf{x},\mathbf{z}}(\mathbf{y}|\mathbf{w} \setminus \mathbf{z})$. We want to use rule 2 in the most efficient way possible and remove as many conditioning variables as we can. The next lemma shows an application of rule 2 on $any$ set does not influence future applications of the rule on other sets elsewhere in the graph.

**Lemma 2** *If rule 2 of $do$-calculus applies to a set **Z** in $G$ for $P_{\mathbf{x}}(\mathbf{y}|\mathbf{w})$ then there are no d-connected paths to **Y** that* *pass through* $\mathbf{Z}$ *in neither* $G_1 = G \setminus \boldsymbol{X}$ *given* $\mathbf{Z}, \boldsymbol{W}$ *nor in* $G_2 = G \setminus (\boldsymbol{X} \cup \boldsymbol{Z})$ *given* $\boldsymbol{W}$*.*

*Proof:* Clearly, there are no d-connected paths through $\mathbf{Z}$ in $G_2$ given $\mathbf{W}$. Consider a d-connected path through $Z \in \mathbf{Z}$ to $\mathbf{Y}$ in $G_1$, given $\mathbf{Z}, \mathbf{W}$. Note that this path must either form a collider at $Z$ or a collider which is an ancestor of $Z$. But this must mean there is a back-door path from $\mathbf{Z}$ to $\mathbf{Y}$, which is impossible, since rule 2 is applicable to $\mathbf{Z}$ in $G$ for $P_{\mathbf{x}}(\mathbf{y}|\mathbf{w})$. Contradiction. □

The following is immediate.

**Corollary 1** *For any* $G$ *and any conditional effect* $P_{\boldsymbol{x}}(\boldsymbol{y}|\boldsymbol{w})$ *there exists a unique maximal set* $\boldsymbol{Z} = \{Z \in \boldsymbol{W} | P_{\boldsymbol{x}}(\boldsymbol{y}|\boldsymbol{w}) = P_{\boldsymbol{x},z}(\boldsymbol{y}|\boldsymbol{w} \setminus \{z\})\}$ *such that rule 2 applies to* $\boldsymbol{Z}$ *in* $G$ *for* $P_{\boldsymbol{x}}(\boldsymbol{y}|\boldsymbol{w})$*.*

*Proof:* Fix two maximal sets $\mathbf{Z}_1, \mathbf{Z}_2 \subseteq \mathbf{W}$ such that rule 2 applies to $\mathbf{Z}_1, \mathbf{Z}_2$ in $G$ for $P_{\mathbf{x}}(\mathbf{y}|\mathbf{w})$. If $\mathbf{Z}_1 \neq \mathbf{Z}_2$, fix $Z \in \mathbf{Z}_1 \setminus \mathbf{Z}_2$. By Lemma 2, rule 2 applies for $\{Z\} \cup \mathbf{Z}_2$ in $G$ for $P_{\mathbf{x}}(\mathbf{y}|\mathbf{w})$, contradicting our assumption.

Thus if we fix $G$ and $P_{\mathbf{x}}(\mathbf{y}|\mathbf{w})$, any set to which rule 2 applies must be a subset of the unique maximal set $\mathbf{Z}$. It follows that $\mathbf{Z} = \{Z \in \mathbf{W} | P_{\mathbf{x}}(\mathbf{y}|\mathbf{w}) = P_{\mathbf{x},z}(\mathbf{y}|\mathbf{w} \setminus \{z\})\}$. □

This corollary states, in particular, that for any $P_{\mathbf{x}}(\mathbf{y}|\mathbf{w})$, we can find a unique maximal set $\mathbf{Z} \subseteq \mathbf{W}$ such that there are no back-door paths from $\mathbf{Z}$ to $\mathbf{Y}$ given the context of the effect, but there is such a back-door path from every $W \in \mathbf{W} \setminus \mathbf{Z}$ to $\mathbf{Y}$.

However, even after a maximal number of nodes is removed from behind the conditioning bar using this corollary, we might still be left with a problem involving conditional distributions. The following key theorem helps us relate this problem to the previously solved case.

**Theorem 6** *Let* $\boldsymbol{Z} \subseteq \boldsymbol{W}$ *be the maximal set such that* $P_{\boldsymbol{x}}(\boldsymbol{y}|\boldsymbol{w}) = P_{\boldsymbol{x},\boldsymbol{z}}(\boldsymbol{y}|\boldsymbol{w} \setminus \boldsymbol{z})$*. Then* $P_{\boldsymbol{x}}(\boldsymbol{y}|\boldsymbol{w})$ *is identifiable in* $G$ *if and only if* $P_{\boldsymbol{x},\boldsymbol{z}}(\boldsymbol{y}, \boldsymbol{w} \setminus \boldsymbol{z})$ *is identifiable in* $G$*.*

*Proof:* See Appendix. □

We can now put the results obtained so far together to construct a simple algorithm for identifiability of conditional effects, shown in Fig. 3.

**Theorem 7 (soundness)** ***IDC*** *is sound.*

*Proof:* The soundness of the first line follows by rule 2 of $do$-calculus. The second line is just a standard conditional distribution calculation, coupled with an invocation of an algorithm known to be sound from [Shpitser & Pearl, 2006]. □

We illustrate the operation of the algorithm by considering the problem of identifying $P_x(y|z)$ in the graph $G$ shown in Fig. 1 (a). Because $Y \perp\!\!\!\perp Z | X, Z$ in $G_{\overline{x},\underline{z}}$, rule 2 applies and

function **IDC**($\mathbf{y}$, $\mathbf{x}$, $\mathbf{z}$, P, G)
INPUT: $\mathbf{x}$,$\mathbf{y}$,$\mathbf{z}$ value assignments, P a probability distribution, G a causal diagram (an I-map of P).
OUTPUT: Expression for $P_{\mathbf{x}}(\mathbf{y}|\mathbf{z})$ in terms of P or **FAIL**(F,F').

1 if $(\exists Z \in \mathbf{Z})(\mathbf{Y} \perp\!\!\!\perp Z | \mathbf{X}, \mathbf{Z} \setminus \{Z\})_{G_{\overline{\mathbf{x}},\underline{z}}}$,
return $\mathbf{IDC}(\mathbf{y}, \mathbf{x} \cup \{z\}, \mathbf{z} \setminus \{z\}, P, G)$.

2 else let $P' = \mathbf{ID}(\mathbf{y} \cup \mathbf{z}, \mathbf{x}, P, G)$.
return $P' / \sum_{\mathbf{y}} P'$.

Figure 3: A complete identification algorithm for conditional effects.

we call the algorithm again with the expression $P_{x,z}(Y)$. This expression is an unconditional effect, so we call **ID** as a subroutine. **ID**, in turn, succeeds immediately on line 6, returning the expression $P(y|x,z)$, which we know is equal to $P(y|z)$ in $G$. Our results so far imply that **IDC** will succeed on all identifiable conditional effects.

**Theorem 8 (completeness)** ***IDC*** *is complete.*

*Proof:* This follows from Theorem 4, Corollary 1 and Theorem 6. □

With a complete algorithm for conditional effects, we can graphically characterize all cases when such effects are identifiable. To do this, we combine Theorem 6 to reduce our problem to one of identifying unconditional effects, and the hedge criterion, which is a complete graphical condition for such effects.

**Corollary 2 (back-door hedge criterion)** *Let* $\boldsymbol{Z} \subseteq \boldsymbol{W}$ *be the unique maximal set such that* $P_{\boldsymbol{x}}(\boldsymbol{y}|\boldsymbol{w}) = P_{\boldsymbol{x},\boldsymbol{z}}(\boldsymbol{y}|\boldsymbol{w} \setminus \boldsymbol{z})$*. Then* $P_{\boldsymbol{x}}(\boldsymbol{y}|\boldsymbol{w})$ *is identifiable from* $P$ *if and only if there does not exist a hedge for* $P_{\boldsymbol{x}'}(\boldsymbol{y}')$*, for any* $\boldsymbol{Y}' \subseteq (\boldsymbol{Y} \cup \boldsymbol{W}) \setminus \boldsymbol{Z}$*,* $\boldsymbol{X}' \subseteq \boldsymbol{X} \cup \boldsymbol{Z}$*.*

*Proof:* This follows from the hedge criterion and Theorem 6. □

The name 'back-door hedge' comes from the fact that both back-door paths and hedge structures are key for identifiability of conditional effects. In particular, $P_{\mathbf{x}}(\mathbf{y}|\mathbf{w})$ is identifiable if and only if $P_{\mathbf{x},\mathbf{z}}(\mathbf{y}, \mathbf{w} \setminus \mathbf{z})$ does not contain any hedges and every $W \in \mathbf{W} \setminus \mathbf{Z}$ has a back-door path to some $Y \in \mathbf{Y}$ in the context of the effect.

## 6 Connections to Existing Algorithms

In this section, we explore the connection of our results to existing identification algorithms for conditional effects. One existing approach to identifying $P_{\mathbf{x}}(\mathbf{y}|\mathbf{w})$ consists of repeatedly using probability manipulations and 3 rules of $do$-calculus until the resulting expression does not involve

function **c-identify**(C, T, Q[T])
INPUT: $T, C \subseteq T$ are both are C-components,
$Q[T]$ a probability distribution
OUTPUT: Expression for $Q[C]$ in terms of $Q[T]$ or **FAIL**

let $A = An(C)_{G_T}$

1 if $A = C$, return $\sum_{T \setminus C} P$

2 if $A = T$, return **FAIL**

3 if $C \subset A \subset T$, there exists a C-component $T'$ such that $C \subset T' \subset A$. return **c-identify**$(C, T', Q[T'])$ ($Q[T']$ is known to be computable from $\sum_{T \setminus A} Q[T]$) any interventional distributions. Our results imply that any identifiable conditional effect can be identified in this way.

Figure 4: A C-component identification algorithm from [Tian, 2004].

**Theorem 9** *The rules of $do$-calculus, together with standard probability manipulations are complete for determining all effects of the form $P_{\mathbf{x}}(\mathbf{y}|\mathbf{z})$.*

*Proof:* The **IDC** algorithm consists of two stages, the first corresponds to repeated applications of rule 2 of $do$-calculus, and the second to identifying the effect of the form $P_{\mathbf{x}}(\mathbf{y})$. The result follows by Theorem 5. □

We next consider an algorithm proposed in [Tian, 2004]. See Fig. 5. This algorithm generalizes an earlier algorithm for unconditional effects [Tian, 2002] which was proven complete in [Shpitser & Pearl, 2006], [Huang & Valtorta, 2006]. Unfortunately, as we now show, the version of the algorithm for conditional effects is not sound.

function **cond-identify**(**y**, **x**, **z**, P, G)
INPUT: **x**,**y**,**z** value assignments, P a probability distribution, G a causal diagram (an I-map of P).
OUTPUT: Expression for $P_{\mathbf{x}}(\mathbf{y}|\mathbf{z})$ in terms of P or **FAIL**.

1 let $D = An(\mathbf{Y} \cup \mathbf{Z})_{G_{\underline{\mathbf{X}}}}, F = D \setminus (\mathbf{Y} \cup \mathbf{Z})$

2 assume $C(D) = \{D_1, ..., D_k\}$

3 let $N = \{D_i | \textbf{c-identify}(D_i, C_{D_i}, Q[C_{D_i}]) = \textbf{FAIL}\}$

4 if $N = \emptyset$, return $\frac{\sum_f \prod_i Q[D_i]}{\sum_{y,f} \prod_i Q[D_i]}$

5 let $F_0 = F \cap (\bigcup_{D_i \in N} Pa(D_i)), I = C(D) \setminus N$

6 remove the set $\{D_i | Pa(D_i) \cap F_0 \neq \emptyset\}$ from $I$ and add it to $I_0$ (which is initially empty)

7 let $B = (F \setminus F_0) \cap \bigcup_{D_i \in I_0} Pa(D_i)$

8 if $B \neq \emptyset$, add all nodes in $B$ to $F_0$, and go to line 6

9 if $\mathbf{Y} \cap (\bigcup_{D_i \in (N \cup I_0)} Pa(D_i)) \neq \emptyset$, return **FAIL**,

else return $\frac{\sum_{f_1} \prod_{D_i \in I_1} Q[D_i]}{\sum_{y,f_1} \prod_{D_i \in I_1} Q[D_i]}$

Figure 5: An identification algorithm from [Tian, 2004]. For each $D_i$, we denote $C_{D_i} \in C(G)$ such that $D_i \subseteq C_{D_i}$.

**Lemma 3** ***cond-identify** is not sound.*

*Proof:* Consider the graph $G''$ in Fig. 1 (c). We will consider the conditional effect $P_x(w|z)$ in this graph. Note that by the back-door hedge criterion this effect is not identifiable in $G''$.

We now trace through the execution of the algorithm for $P_x(w|z)$. In this case, $D = \{Y, Z, W\}, F = \{Y\}, C(G) = \{\{X, Z\}, \{Y\}\{W\}\}, C(D) = \{\{Z\}, \{Y\}, \{W\}\}$. Now identification of $Q[\{Y\}]$ and $Q[\{W\}]$ trivially succeeds, while identification of $Q[\{Z\}]$ from $Q[\{Z, X\}]$ fails. Therefore, $I = \{\{Y\}, \{W\}\}, N = \{\{Z\}\}, F_0 = F$. Because $\{Y\}$ is not a parent of any identifiable C-component, line 6 does nothing. Because $F_0 = F$, line 8 does nothing. However, $\{W\} \cap \{Y\} = \emptyset$, so the algorithm succeeds. This implies the result. □

## 7 Conclusions

We have presented a complete graphical criterion for identification of conditional interventional distributions in semi-Markovian causal models. We used this criterion to construct a sound and complete identification algorithm for such distributions, and prove completeness of a $do$-calculus for the same identification problem.

This work closes long standing questions about identifiability of interventional distributions, but much more remains to be done. Certain kinds of causal effects or counterfactual statements of interest cannot be expressed as an interventional distribution. For instance, certain kinds of direct and indirect effects [Pearl, 2001], and path-specific effects are represented instead as a probability of a formula in a certain modal logic [Avin, Shpitser, & Pearl, 2005]. Questions about identifiability of such effects in semi-Markovian models are an open problem for future work.

**Acknowledgments**

The authors thank anonymous reviewers for helpful comments. This work was supported in part by AFOSR grant #F49620-01-1-0055, NSF grant #IIS-0535223, and MURI grant #N00014-00-1-0617.

## Appendix

We first prove two utility lemmas.

**Lemma 4** *Let $M$ be a causal model. Assume $P(\mathbf{y}) > 0$. Then for any $\mathbf{X}$ disjoint from $\mathbf{Y}$, there exists $\mathbf{x}$ such that $P_{\mathbf{x}}(\mathbf{y}) > 0$.*

*Proof:* Let $\mathbf{U}$ be the set of unobservable variables in $M$. We know that $P(\mathbf{y}) = \sum_{\mathbf{Y(u)=y}} P(\mathbf{u})$. Fix $\mathbf{u}$ such that $\mathbf{Y}(\mathbf{u}) = \mathbf{y}$. We know such a $\mathbf{u}$ exists because $P(\mathbf{y}) > 0$. We also know $\mathbf{u}$ renders $M$ deterministic. Let $\mathbf{x}$ be the value $\mathbf{X}(\mathbf{u})$ assumes. Our conclusion now follows. □

**Lemma 5** *Let $F, F'$ form a hedge for $P_{\mathbf{x}}(\mathbf{y})$. Then $F \subseteq F' \cup \mathbf{X}$.*

*Proof:* It has been shown that **ID** fails on $P_{\mathbf{x}}(\mathbf{y})$ in $G$ and returns a hedge if and only if $P_{\mathbf{x}}(\mathbf{y})$ is not identifiable in $G$. In particular, edge subgraphs of the graphs $G$ and $S$ returned by line 5 of **ID** form the C-forests of the hedge in question. It is easy to check that a subset of $\mathbf{X}$ and $S$ partition $G$. □

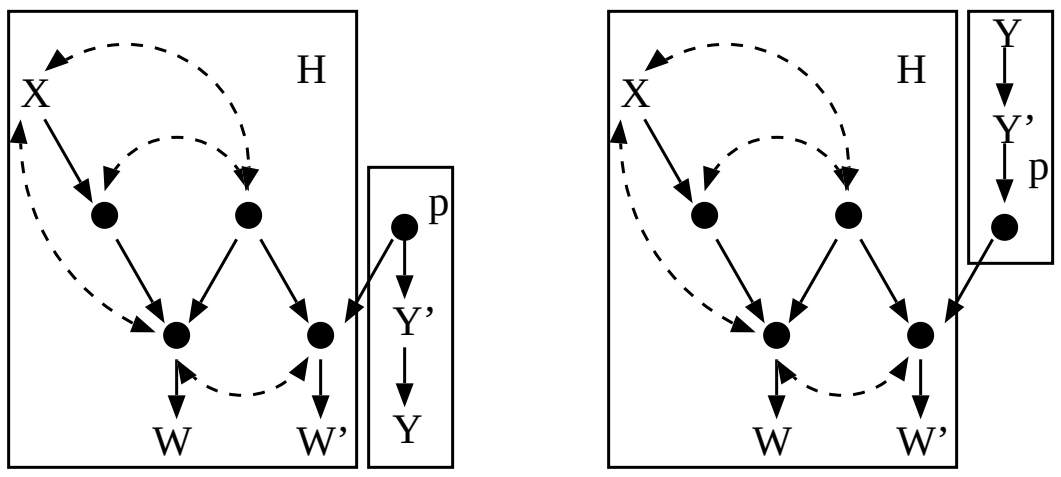

Figure 6: Inductive cases for proving non-identifiability of $P_x(y|w, w')$.

Next, we rephrase the statement of the theorem slightly to reduce 'algebraic clutter.'

**Theorem 6** *Let $P_{\mathbf{x}}(\mathbf{y}|\mathbf{w})$ be such that every $W \in \mathbf{W}$ has a back-door path to $\mathbf{Y}$ in $G \setminus \mathbf{X}$ given $\mathbf{W} \setminus \{W\}$. Then $P_{\mathbf{x}}(\mathbf{y}|\mathbf{w})$ is identifiable in $G$ if and only if $P_{\mathbf{x}}(\mathbf{y}, \mathbf{w})$ is identifiable in $G$.*

*Proof:* If $P_{\mathbf{x}}(\mathbf{y}, \mathbf{w})$ is identifiable in $G$, then we can certainly identify $P_{\mathbf{x}}(\mathbf{y}|\mathbf{w})$ by marginalization and division. The difficult part is to prove that if $P_{\mathbf{x}}(\mathbf{y}, \mathbf{w})$ is not identifiable then neither is $P_{\mathbf{x}}(\mathbf{y}|\mathbf{w})$.

Assume $P_{\mathbf{x}}(\mathbf{w})$ is identifiable. Then if $P_{\mathbf{x}}(\mathbf{y}|\mathbf{w})$ were identifiable, we would be able to compute $P_{\mathbf{x}}(\mathbf{y}, \mathbf{w})$ by the chain rule. Thus our conclusion follows.

Assume $P_{\mathbf{x}}(\mathbf{w})$ is not identifiable. We also know that every $W \in \mathbf{W}$ contains a back-door path to some $Y \in \mathbf{Y}$ in $G \setminus \mathbf{X}$ given $\mathbf{W} \setminus \{W\}$. Fix such $W$ and $Y$, along with a subgraph $p$ of $G$ which forms the witnessing back-door path. Consider also the hedge $F, F'$ which witnesses the non-identifiability of $P_{\mathbf{x}'}(\mathbf{w}')$, where $\mathbf{X}' \subseteq \mathbf{X}, \mathbf{W}' \subseteq \mathbf{W}$.

Let $H = De(F)_G \cup An(\mathbf{W}')_G$. We will attempt to show that $P_{\mathbf{x}'}(Y|\mathbf{w})$ is not identifiable in $H \cup p$. Without loss of generality, we make the following three assumptions. First, we restrict our attention to $\mathbf{W}'' \subseteq \mathbf{W}$ that occurs in $H \cup p$. Second, we assume all observable nodes in $H$ have at most one child. Finally, we assume $p$ is a path segment which starts at $H$ and ends at $Y$, and does not intersect $H$. This assumes $Y \notin H$. We will handle the case when $Y \in H$ in one of the base cases.

Consider the models $M^1, M^2$ from the proof of Theorem 2 which induce $H$. We extend the models by adding to them binary variables in $p$. Each variable $X \in p$ is set to the bit parity of its parents, if it has any. If not, $X$ behaves as a fair coin. If $Y \in H$ has a parent $X \in p$, the value of $Y$ is set to the bit parity of all parents of $Y$, including $X$.

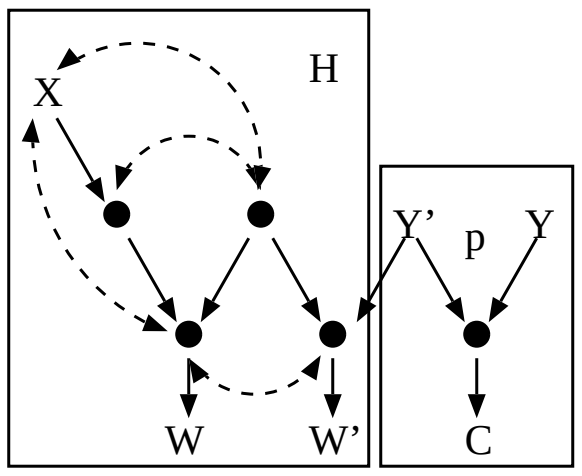

(a) (b)

Figure 7: Inductive cases for proving non-identifiability of $P_x(y|w, w')$.

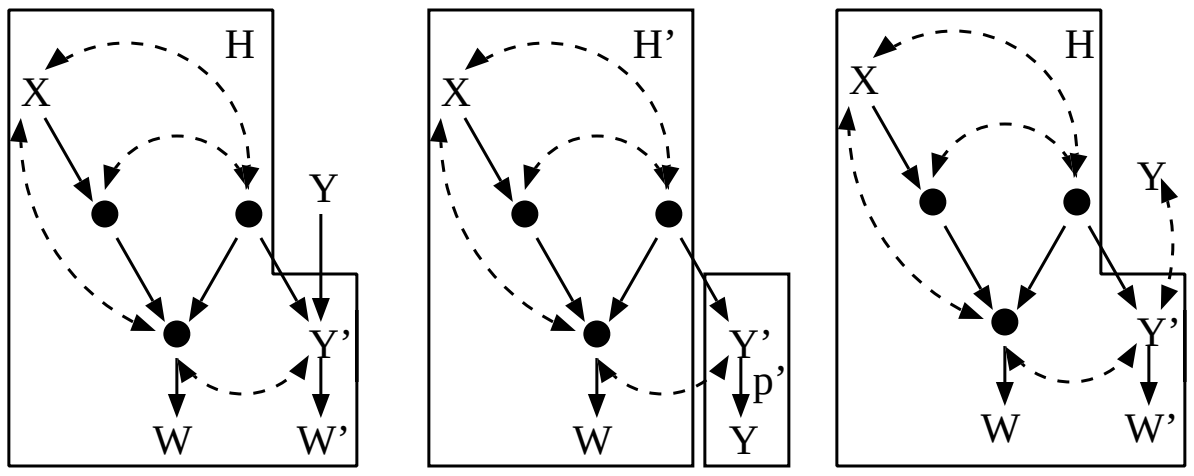

(a) (b) (c)

Figure 8: Base cases for proving non-identifiability of $P_x(y|w, w')$.

Call the resulting models $M^1_*, M^2_*$. Because $M^1, M^2$ agreed on $P(H)$, and variables and functions in $p$ are the same in both models, $P^1_* = P^2_*$. It has already been shown that $P^1(\mathbf{x}|Pa(\mathbf{X})_G \setminus \mathbf{X}) > 0$, which implies the same is true for $P^1_*$. We will assume $\mathbf{w}''$ assigns 0 to every variable in $\mathbf{W}''$. Note that in $M^1_*$, $\mathbf{w}''$ is equal to the bit parity of all the $\mathbf{U}$ nodes in $H$ and all parent-less nodes in $p$. Similarly, in $M^2_*$ $\mathbf{w}''$ is equal to the bit parity of all the $\mathbf{U}$ nodes in $F'$ and all parent-less nodes in $p$. It's easy to see that $P^1_*(\mathbf{w}'') > 0$ and $P^2_*(\mathbf{w}'') > 0$. Now by Lemma 4, $P^1_{*\mathbf{x}}(\mathbf{w}'') > 0$ and $P^2_{*\mathbf{x}}(\mathbf{w}'') > 0$.

What remains to be shown is that $P^1_{*\mathbf{x}}(y|\mathbf{w}'') \neq P^2_{*\mathbf{x}}(y|\mathbf{w}'')$. We will prove this by induction on the path structure of $p$. We handle the inductive cases first. In all these cases, we fix a node $Y'$ that is between $Y$ and $H$ on the path $p$, and prove that if $P_{\mathbf{x}'}(y'|\mathbf{w}'')$ is not identifiable, then neither is $P_{\mathbf{x}'}(y|\mathbf{w}'')$. Note that despite the fact that we represent variable marginalization as a multiplication by a matrix as a matter of convenience, we make sure to only sum over values with positive probability of occurrence in the given context.

Assume neither $Y$ nor $Y'$ have descendants in $\mathbf{W}''$. If $Y'$ is a parent of $Y$ as in Fig. 6 (a), then $P_{\mathbf{x}'}(y|\mathbf{w}'') = \sum_{y'} P(y|y')P_{\mathbf{x}'}(y'|\mathbf{w}'')$. If $Y$ is a parent of $Y'$, as in Fig. 6 (b) then the next node in $p$ must be a child of $Y'$. Therefore, $P_{\mathbf{x}'}(y|\mathbf{w}'') = \sum_{y'} P(y|y')P_{\mathbf{x}'}(y'|\mathbf{w}'')$. In either case, by construction $P(Y|Y')$ is a 2 by 2 identity matrix. This implies that the mapping from $P_{\mathbf{x}'}(y'|\mathbf{w}'')$ to $P_{\mathbf{x}'}(y|\mathbf{w}'')$ is one to one. If $Y'$ and $Y$ share a hidden common parent $U$ as in Fig. 7 (b), then our result follows by combining the previous two cases.

The next case is if $Y$ and $Y$ have a common child $C$ which is either in $\mathbf{W}''$ or has a descendant in $\mathbf{W}''$, as in Fig. 7 (a). Now $P_{\mathbf{x}'}(y|\mathbf{w}'') = \sum_{y'} P(y|y', c)P_{\mathbf{x}'}(y'|\mathbf{w}'')$. Because all nodes in $\mathbf{W}''$ were observed to be 0, $P(y|y', c)$ is again a 2 by 2 identity matrix.

Finally, we handle the base cases of our induction. In all such cases, $Y$ is the first node not in $H$ on the path $p$. Let $Y'$ be the last node in $H$ on the path $p$.

Assume $Y$ is a parent of $Y'$, as shown in Fig. 8 (a). By Lemma 5, we can assume $Y \notin An(F \setminus F')_H$. By construction, $(\sum \mathbf{W}'' = Y + 2 * \sum \mathbf{U}) \pmod 2$ in $M^1_*$, and $(\sum \mathbf{W}'' = Y + 2 * \sum(\mathbf{U} \cap F')) \pmod 2$ in $M^2_*$. If every variable in $\mathbf{W}''$ is observed to be 0, then $Y = (2 * \sum \mathbf{U}) \pmod 2$ in $M^1_*$, and $Y = (2 * \sum(\mathbf{U} \cap F')) \pmod 2$ in $M^2_*$. If an intervention $do(\mathbf{x})$ is performed, $(\sum \mathbf{W}'' = Y + 2 * \sum(\mathbf{U} \cap F')) \pmod 2$ in $M^2_{*\mathbf{x}}$, by construction. Thus if $\mathbf{W}''$ are all observed to be zero, $Y = 0$ with probability 1. It was shown in [Shpitser & Pearl, 2006] that in $M^1_{\mathbf{x}}$, $(\sum \mathbf{w}'' = \mathbf{x} + \sum \mathbf{U}') \pmod 2$, where $\mathbf{U}' \subseteq \mathbf{U}$ consists of unobservable nodes with one child in $An(\mathbf{X})_F$ and one child in $F \setminus An(\mathbf{X})_F$. Because $Y \notin An(F \setminus F')_H$, we can conclude that if $\mathbf{W}''$ are observed to be 0, $Y = (\mathbf{x} + \sum \mathbf{U}') \pmod 2$ in $M^1_{*\mathbf{x}'}$. Thus, $Y = 0$ with probability less than 1. Therefore, $P^1_{*\mathbf{x}'}(y|\mathbf{w}'') \neq P^2_{*\mathbf{x}'}(y|\mathbf{w}'')$ in this case.

Assume $Y$ is a child of $Y'$. Now consider a graph $G'$ which is obtained from $H \cup p$ by removing the (unique) outgoing arrow from $Y'$ in $H$. If $P_{\mathbf{x}'}(y|\mathbf{w}'')$ is not identifiable in $G'$, we are done. Assume $P_{\mathbf{x}'}(y|\mathbf{w}'')$ is identifiable in $G'$. If $Y' \in F$, and $\mathbf{R}$ is the root set of $F$, then removing the $Y'$-outgoing directed arrow from $F$ results in a new C-forest, with a root set $\mathbf{R} \cup \{Y'\}$. Because $Y$ is a child of $Y'$, the new C-forests form a hedge for $P_{\mathbf{x}'}(y, \mathbf{w}'')$. If $Y' \in H \setminus F$, then removing the $Y'$-outgoing directed arrow results in substituting $Y$ for $W \in \mathbf{W}'' \cap De(Y')_H$. Thus in $G'$, $F, F'$ form a hedge for $P_{\mathbf{x}'}(y, \mathbf{w}'' \setminus \{w\})$. In either case, $P_{\mathbf{x}'}(y, \mathbf{w}'')$ is not identifiable in $G'$.

Now if $P_{\mathbf{x}'}(\mathbf{w}'')$ is identifiable in $G'$, we are done. If not, consider a smaller hedge $H' \subset H$ witnessing this fact. Now consider the segment $p'$ of $p$ between $Y$ and $H'$. We can repeat the inductive argument for $H'$, $p'$ and $Y$. See Fig. 8 (b). Note that this base case also handles the case when $Y \in H$. We just let $Y = Y'$, and apply the previous reasoning.

If $Y$ and $Y'$ have a hidden common parent, as is the case in Fig. 8 (c), we can combine the first inductive case, and the first base case to prove our result.

This completes the proof. □